%% file: root.tex
\documentclass[letterpaper, 10 pt, conference]{ieeeconf}  
\IEEEoverridecommandlockouts                             
\usepackage{cite}
\usepackage{amsmath,amssymb,amsfonts}
\usepackage{algorithmic} 
\usepackage{graphicx}
\usepackage{textcomp}
\usepackage{tabularx}
\usepackage{siunitx}
\usepackage{xcolor}
\usepackage{balance}
\usepackage{bm}
\usepackage[ruled,linesnumbered]{algorithm2e}
\usepackage{subcaption} 
\usepackage[font={small}]{caption}
\usepackage{multirow}
\usepackage[colorlinks, linkcolor=black, anchorcolor=black, citecolor=black]{hyperref}
\usepackage{threeparttable}
\usepackage{mathtools}
\usepackage{booktabs}
\usepackage{tablefootnote}
\usepackage{pifont}
\usepackage{anyfontsize}
\usepackage{wrapfig}
\newcommand{\cmark}{\ding{51}}  % 对号
\newcommand{\xmark}{\ding{55}}  % 叉号

\title{\LARGE \bf
LADY: \underline{L}inear \underline{A}ttention for Autonomous \underline{D}riving Efficienc\underline{y} without Transformers
}

\author{Jihao Huang$^{1*}$, Xi Xia$^{1*}$, Zhiyuan Li$^{3}$, Tianle Liu$^{1,2}$, Jingke Wang$^{1}$, Junbo Chen$^{1}$, Tengju Ye$^{1\dagger}$
\thanks{* Equal Contributions.}
\thanks{$^{1}$ Udeer AI, Hangzhou, China.
$^{2}$ Zhejiang University, Hangzhou, China.
$^{3}$ Yuanshi Intelligence, Shenzhen, China. ($^\dagger$Corresponding author.)
}
}

\begin{document}
\maketitle
\thispagestyle{empty}
\pagestyle{empty}

%%%%%%%%%%%%
% abstract %
%%%%%%%%%%%%
\input{sections/abstract}

%%%%%%%%%%%%
% sections %
%%%%%%%%%%%%
\input{sections/introduction}
\input{sections/background}
\input{sections/method}
\input{sections/exp}
\input{sections/conclusion}

{
\bibliographystyle{IEEEtran}
\bibliography{ref}
% \balance
}

\end{document}

%% file: sections/abstract.tex
\begin{abstract}
End-to-end autonomous driving has emerged as a promising paradigm.
However, state-of-the-art methods rely heavily on Transformer architectures. 
The inherent quadratic complexity of Transformers restricts their ability to model long-range spatial and temporal dependencies, particularly on resource-constrained edge platforms. 
Given the inherent demand for efficient temporal modeling in autonomous driving, this computational bottleneck severely constrains real-time deployment.
While linear attention mechanisms offer a computationally efficient alternative, existing architectures are predominantly limited to self-attention, lacking the cross-modal capabilities essential for autonomous driving.
In this work, we propose LADY, the first fully linear attention-based generative model for end-to-end autonomous driving. 
LADY incorporates a novel, lightweight linear cross-attention (LICA) mechanism to enable effective cross-modal interaction while preserving linearity. 
A key advantage of our framework is its ability to fuse long-range temporal contexts during inference with constant computational and memory costs ($O(1)$), regardless of the historical sequence length. 
Experiments on the NAVSIM and Bench2Drive benchmarks demonstrate that LADY achieves performance comparable to state-of-the-art methods, delivering competitive planning accuracy with significantly reduced latency.
Furthermore, efficiency benchmarking on edge devices validates the model's feasibility for resource-limited scenarios.
\end{abstract}

%% file: sections/introduction.tex
\section{Introduction}
With the rapid evolution of artificial intelligence, autonomous driving has garnered widespread attention from both academia and industry, promising safer roads and more efficient transportation~\cite{li2024end,hu2023planning, zheng2025genad}. 
While impressive progress has been made in localization and scene perception~\cite{zhao2024survey,lu2022real,wang2023unitr,chen2026video}, we focus on the critical challenge of motion planning~\cite{pan2024safe,wang2024llm}, the robustness of which directly determines ride safety. 
The primary objective of motion planning is to generate collision-free, smooth, and dynamically feasible trajectories in real time~\cite{zhao2024survey}. 
Traditional rule-based planners, relying on hand-crafted heuristics, often struggle with corner cases and become brittle when facing the stochastic nature of real-world traffic~\cite{teng2023motion}. 
In contrast, end-to-end learning-based planners map raw sensor inputs directly to planned trajectories, offering superior scalability and robustness against complex scenarios~\cite{chib2023recent,li2024ego}.

\begin{figure}[!t]
    \centering
    \includegraphics[width=1.0\linewidth]{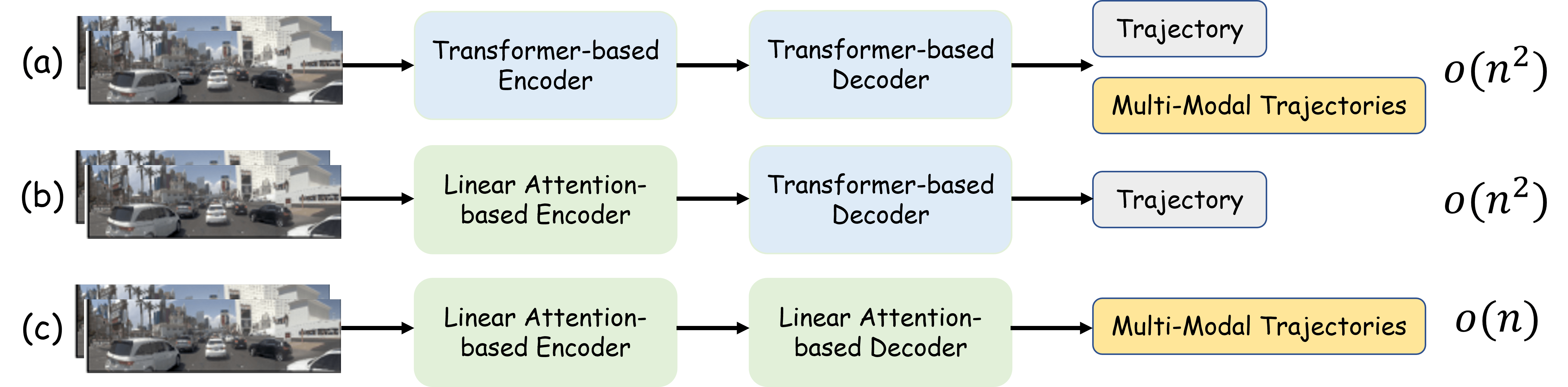}
    \caption{Comparison of different end-to-end paradigms: (a) - (b) Existing methods relying on computationally expensive Transformer cross-attention versus (c) our proposed LADY, which is constructed entirely using lightweight linear attention mechanisms.}
    \label{fig:overall_fig}
\end{figure}

Dominant end-to-end methods—such as UniAD~\cite{hu2023planning}, Transfuser~\cite{chitta2022transfuser}, FusionAD~\cite{ye2023fusionad}, and VAD~\cite{jiang2023vad}—typically imitate expert drivers to output a single deterministic trajectory, as shown in Fig.~\ref{fig:overall_fig}(a).
However, human driving behavior is inherently multi-modal: in any given scenario, multiple plausible maneuvers may exist (e.g., yielding vs. overtaking). 
To address this, recent approaches have shifted towards generating multi-modal trajectory candidates to capture diverse driving intentions, as shown in Fig.~\ref{fig:overall_fig}(a).
For instance, VADv2~\cite{chen2024vadv2} and the Hydra-MDP series~\cite{li2024hydra,li2025hydra} sample from a large, fixed vocabulary of anchor trajectories to generate multiple feasible driving maneuvers. 
Inspired by the diffusion policy~\cite{chi2023diffusion}, DiffusionDrive~\cite{liao2025diffusiondrive} introduces a truncated diffusion process to end-to-end driving, mitigating mode collapse and reducing computational overhead while yielding diverse, high-quality candidates. 
Despite these advances, most existing works rely heavily on Transformer architectures for both feature fusion and cross-attention. 
This imposes a significant computational burden, limiting their deployment in real-time, resource-constrained applications.

To mitigate computational costs, recent works like DRAMA~\cite{yuan2024drama} have proposed Mamba-based encoder-decoders for efficient sensor fusion, as illustrated in Fig.~\ref{fig:overall_fig}(b). 
However, DRAMA still retains a Transformer-based module for cross-attention due to mismatched query and feature lengths, thereby failing to eliminate the quadratic cost of softmax attention. 
Furthermore, the aforementioned approaches typically rely solely on the camera and LiDAR features from the most recent frame, neglecting valuable temporal context from preceding frames. 
In dynamic scenarios—such as a pedestrian emerging from behind an occlusion or a cyclist accelerating through an intersection—single-frame information is often insufficient to anticipate motion patterns, leading to unsafe or overly conservative plans. While integrating multi-frame history is essential for robust prediction, applying Transformer-based backbones to process long temporal sequences incurs prohibitively high computation due to the quadratic complexity of attention mechanisms. 
Therefore, there is an urgent need for an end-to-end framework capable of efficiently fusing multi-frame sensor information without compromising real-time performance on edge devices.

In the realm of sequence modeling, linear attention mechanisms~\cite{gu2023mamba,dao2024transformers,peng2023rwkv,peng2024eagle,peng2025rwkv,kimiteam2025kda} have gained widespread attention for combining the parallel training efficiency of Transformers with the lightweight, constant-time inference of RNNs. 
Building upon the latest RWKV-7 architecture, we propose LADY, a fully linear attention-based framework designed to efficiently fuse multi-frame camera and LiDAR features for multi-modal trajectory generation, as depicted in Fig.~\ref{fig:overall_fig}(c).
By aiming to leverage the efficiency of linear attention, LADY enhances planning performance while substantially reducing computational overhead.
Notably, our framework is designed with modular extensibility, allowing for seamless adaptation to other linear attention variants beyond RWKV-7, thus offering flexibility for future optimization.

Our main contributions are summarized as follows:
\begin{itemize}
    \item To the best of our knowledge, LADY is the first fully linear attention-based end-to-end autonomous driving model, eliminating quadratic complexity components throughout the pipeline.
    \item We propose a lightweight linear cross-attention (LICA) mechanism to ensure end-to-end linearity. 
    Integrated with a diffusion-based trajectory decoder, it enhances the diversity and quality of multi-modal trajectories.
    \item LADY is capable of fusing long-range temporal context during inference with constant computational and memory overhead ($O(1)$ complexity), regardless of the historical sequence length, thereby significantly enhancing planning robustness.
    \item Experimental results demonstrate that LADY achieves superior trade-offs between planning accuracy and runtime efficiency on the NAVSIM and Bench2Drive benchmarks, with edge validation confirming its practical applicability for real-world deployment.
\end{itemize}

%% file: sections/background.tex
\section{Related Works}
\subsection{End-to-End Autonomous Driving}
Leveraging rapid advancements in perception algorithms and computational hardware, end-to-end autonomous driving has emerged as a promising paradigm, mapping raw sensor inputs directly to feasible trajectories.
Following the taxonomy established in~\cite{jiang2025transdiffuser}, contemporary approaches can be categorized into three primary streams: regression-based, scoring-based, and diffusion-based paradigms.

\textbf{Regression-based approaches}~\cite{chitta2022transfuser,hu2023planning,ye2023fusionad,jiang2023vad,yuan2024drama} typically predict a single deterministic trajectory in an auto-regressive manner.
UniAD~\cite{hu2023planning} unifies multiple perception tasks—including object detection and semantic segmentation—into a unified planning network, demonstrating the scalability of end-to-end systems.
Similarly, Transfuser~\cite{chitta2022transfuser} employs a multi-scale fusion of camera and LiDAR features to regress waypoints. 
It has served as a standard baseline on the NAVSIM benchmark~\cite{dauner2024navsim}, which aligns open-loop and closed-loop evaluation metrics and provides a public evaluation server for end-to-end planning.

\textbf{Scoring-based approaches}~\cite{li2024hydra,chen2024vadv2,li2025hydra,guo2025ipad} address the multi-modal nature of human driving by generating diverse trajectory candidates and selecting the optimal one via scoring functions.
VADv2~\cite{chen2024vadv2} extends the regression-based VAD~\cite{jiang2023vad} by sampling from a large, fixed vocabulary of anchor paths, selecting the best maneuver based on probabilistic scores.
The Hydra-MDP series~\cite{li2024hydra,li2025hydra} further refines this mechanism by incorporating rule-based teacher supervision, allowing the model to distill expert heuristics while learning human-like behaviors.
Additionally, iPad~\cite{guo2025ipad} introduces a dedicated scorer to explicitly evaluate the safety, efficiency, and comfort of proposals, leveraging ground-truth metrics derived from NAVSIM~\cite{dauner2024navsim} for supervision.

\textbf{Diffusion-based approaches}~\cite{liao2025diffusiondrive,jiang2025transdiffuser} model trajectory generation as a conditional denoising process, inspired by the success of generative diffusion models.
DiffusionDrive~\cite{liao2025diffusiondrive} pioneers this direction by introducing a truncated diffusion policy, effectively mitigating mode collapse and reducing inference latency compared to standard diffusion pipelines.
Building on this, TransDiffuser~\cite{jiang2025transdiffuser} incorporates a representation decorrelation mechanism during training, further enhancing the diversity and kinematic quality of the generated multi-modal trajectories.

\subsection{Application of Linear Attention in Autonomous Driving}
\begin{figure*}[!t]
    \centering
    \includegraphics[width=\linewidth]{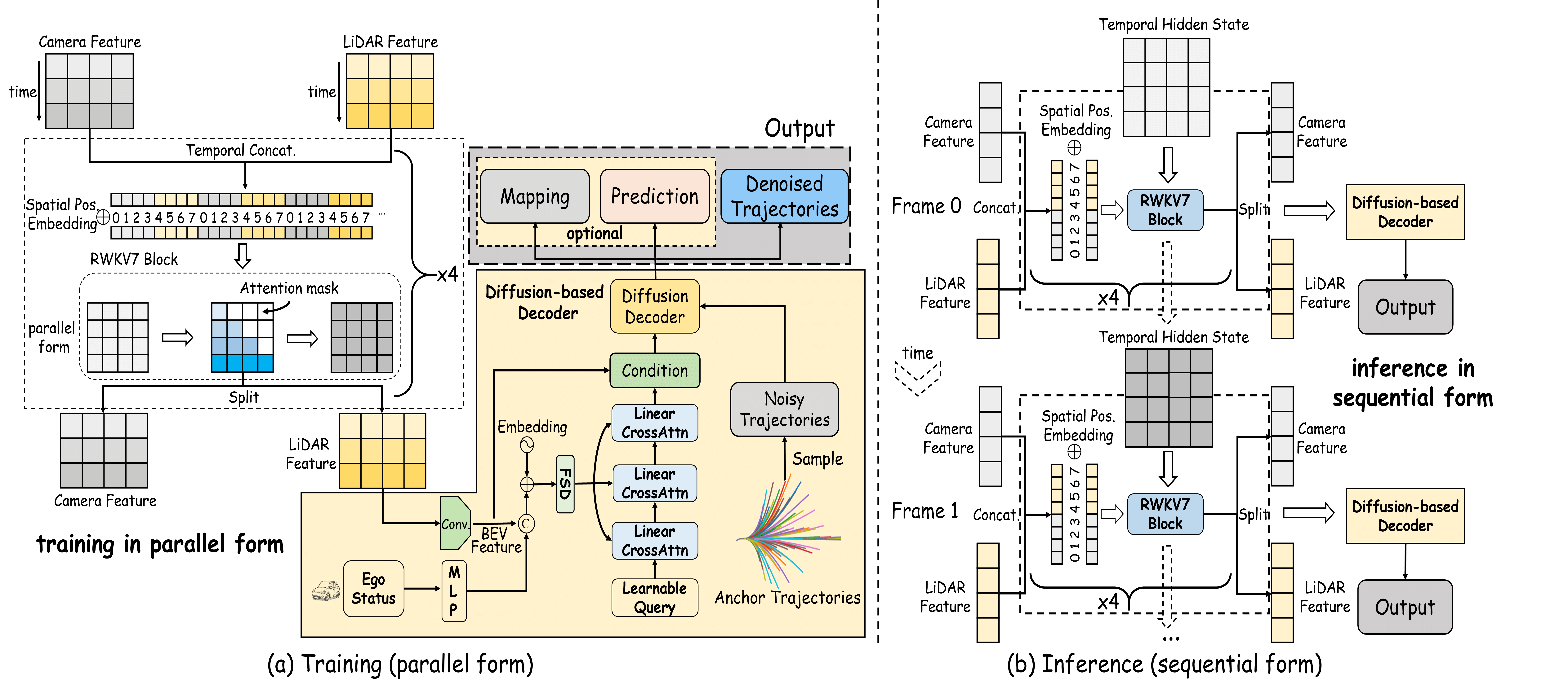}
    \caption{Overall architecture of LADY. The framework takes the ego vehicle's status, historical multi-modal sensor streams (camera and LiDAR), and noisy anchor trajectories as inputs. It outputs denoised multi-modal trajectories, passability maps, and collision-aware agent future states. During training, the model leverages linear attention to fuse multi-frame features in \textbf{parallel}. In contrast, during inference, it operates in a \textbf{recurrent} mode, sequentially updating the temporal hidden state with current frame features to generate BEV feature.}
    \label{fig:whole}
\end{figure*}
Despite the success of Transformer-based architectures in capturing global context, their inherent quadratic computational complexity ($O(N^2)$) regarding sequence length poses a significant bottleneck for real-time autonomous driving.
To address this, linear attention architectures—such as RWKV~\cite{peng2024eagle,peng2025rwkv}, Gated Linear Attention~\cite{yang2024gated}, and Mamba-2~\cite{dao2024transformers}, and KDA~\cite{kimiteam2025kda}—reformulate attention as a RNN. 
This formulation enables parallel training while allowing for constant-time ($O(1)$) inference and constant memory usage.

DRAMA~\cite{yuan2024drama} represents an early attempt to apply Mamba-2 for fusing camera and LiDAR features, achieving improved encoding efficiency.
However, DRAMA adopts a hybrid architecture: while it uses Mamba for self-attention within each individual modality, it still relies on standard Transformer-based mechanisms for cross-modal fusion due to the challenge of aligning variable-length sequences.
Consequently, DRAMA retains quadratic complexity bottlenecks in its decoding stage and cannot be considered a fully linear framework.
Moreover, its regression-based design limits its ability to model complex multi-modal distributions, preventing it from achieving state-of-the-art (SOTA) performance.

Beyond architectural hybridity, the underlying state-space mechanism also differentiates current approaches.
While Mamba-2 improves efficiency, its state update rule primarily relies on a decay mechanism, which suppresses older information rather than explicitly removing it. 
This can lead to historical data lingering as a diminishing fraction of the state.
In contrast, DeltaNet~\cite{schlag2021linear} applies the Error Correcting Delta Rule~\cite{widrow1988adaptive} to update key-value states, allowing for the explicit addition and removal of information.
Building on this, RWKV-7~\cite{peng2025rwkv} generalizes the delta rule with parallelizable diagonal-plus-low-rank updates~\cite{yang2024parallelizing}, overcoming the retention limitations of Mamba-2 while preserving linear complexity, enabling efficient training of models.

Crucially, most existing end-to-end planners utilize only the most recent sensor frame, lacking the \textbf{temporal context} necessary to anticipate dynamic agents and occluded hazards.
While incorporating multi-frame sensor information is vital, doing so with Transformers incurs prohibitive computational costs.
To solve this dilemma, we propose leveraging the advanced RWKV-7 backbone to fuse multi-frame camera and LiDAR features.
Furthermore, to ensure end-to-end linearity, we introduce a lightweight \textbf{LICA} mechanism that aligns modalities with linear time and space complexity.
By integrating these components, LADY establishes the first fully linear attention-based architecture for end-to-end autonomous driving, enabling robust long-term temporal modeling without the computational penalty of Transformers.

%% file: sections/method.tex
\section{Method Design}
The overall architecture of LADY adopts an encoder-decoder framework, as illustrated in Fig.~\ref{fig:whole}.
The encoder integrates features from multi-frame camera and LiDAR inputs to construct Bird's-Eye-View (BEV) representations, which are subsequently decoded to predict multi-modal trajectories. 
Notably, LADY employs a dual-mode temporal strategy: during training, it processes multi-frame features in parallel to maximize computational efficiency, as shown in Fig.~\ref{fig:whole}(a).
Conversely, during inference, it fuses the current frame camera and LiDAR features with the temporal hidden state sequentially, as shown in Fig.~\ref{fig:whole}(b).
This design enables the effective aggregation of long-range temporal contexts in real-time. 
In the following, we provide a detailed description of the key components of LADY.

\subsection{Linear Attention-based Multi-Frame Feature Fusion}
We employ the ResNet-34~\cite{he2016deep} backbone, following Transfuser~\cite{chitta2022transfuser}, to extract BEV feature maps from multi-frame camera and LiDAR inputs. 
To enable efficient processing of long-range historical information, we adopt the off-the-shelf RWKV-7 architecture~\cite{peng2025rwkv} as our temporal fusion engine. 
Unlike standard Transformers with quadratic complexity, RWKV-7 utilizes a linear attention mechanism, allowing it to function as a RNN during inference. 
Its core recurrent state update is formulated as:
\begin{equation*}
    \bm{S}_t = \bm{S}_{t - 1} + \bm{k}_t \bm{v}_t^\top, \quad \bm{o}_t = \bm{S}_t \bm{q}_t,
\end{equation*}
where $\bm{S}_t$ represents the time-decaying hidden state. 
This formulation ensures linear time complexity and constant memory usage during inference. 
For complete mathematical derivations and architectural details of the RWKV-7 block, we refer readers to~\cite{peng2025rwkv}.

The primary novelty of LADY lies in how we adapt this linear attention mechanism for multi-modal sensor fusion. 
Specifically, during training, the camera and LiDAR features are organized into separate sequences across different time steps, and we construct a unified sequence by concatenating the camera and LiDAR features at each time step, augmented with learnable spatial positional embeddings, as shown in Fig.~\ref{fig:whole}(a).
These concatenated multi-frame features are fed into stacked RWKV-7 blocks, enabling simultaneous interaction across both temporal and cross-modal dimensions. 
After fusion, the sequence is split back into individual sensor streams.

A key advantage of our design is the discrepancy between training and inference stages. 
While training exploits parallel computation, the inference stage operates sequentially: it integrates current sensor features conditioned on the historical state $\bm{S}_{t-1}$, as shown in Fig.~\ref{fig:whole}(b).
This allows LADY to incorporate extensive temporal context without the computational overhead typical of sliding-window Transformers, which is critical for autonomous driving.
Following multi-resolution feature fusion, we apply convolutions to the fused LiDAR features to generate the BEV feature map, which is then concatenated with the ego-status and added learnable positional embeddings to construct a context-rich representation for downstream tasks.
Finally, to improve model robustness, we employ the feature state dropout (FSD)~\cite{yuan2024drama}, which stochastically drops either the BEV features or ego-state channels with distinct probabilities.

\subsection{Linear Cross-Attention Mechanism}
\begin{figure}[!t]
    \centering
    \includegraphics[width=0.9\linewidth]{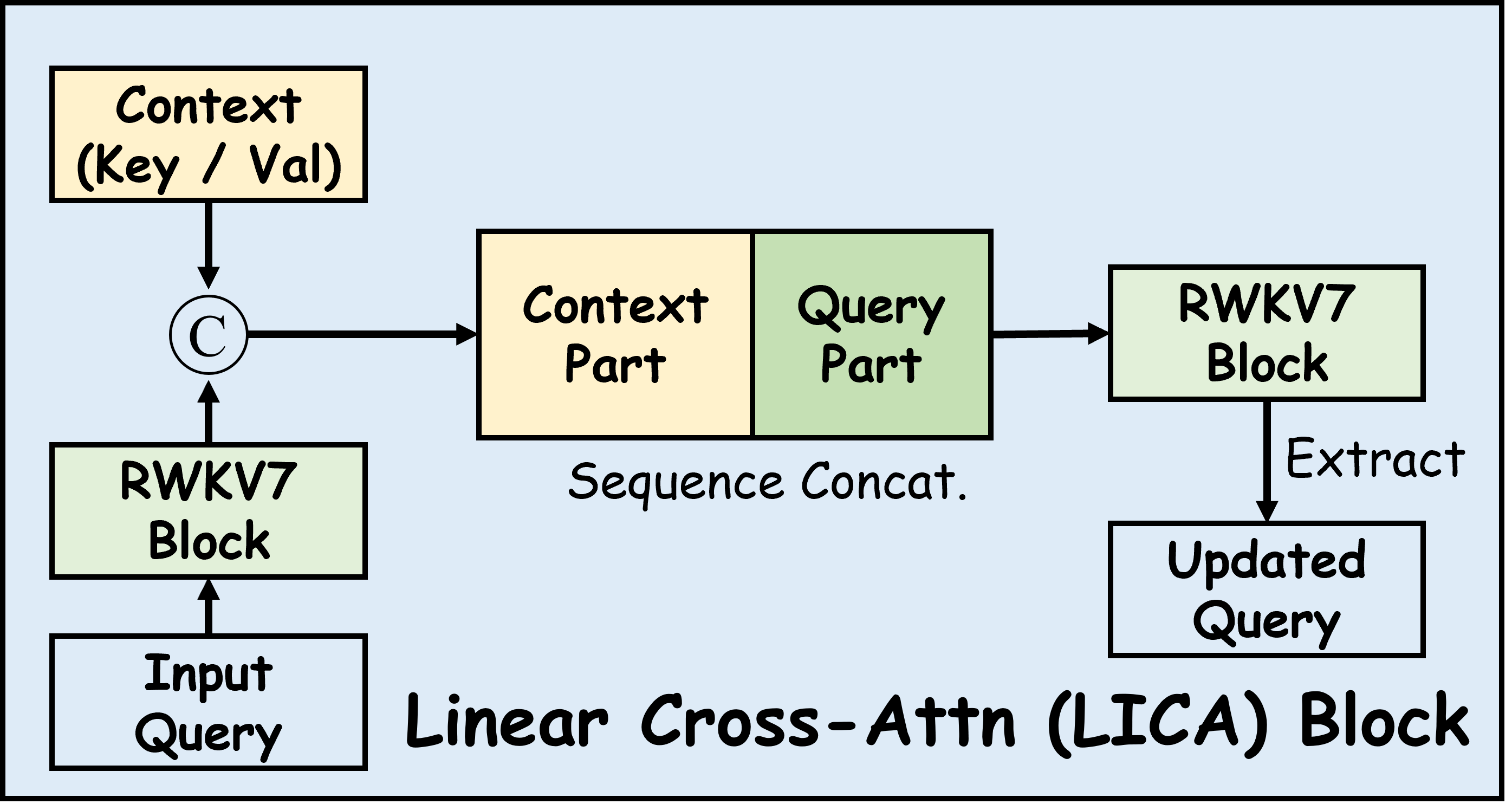}
    \caption{Overall architecture of linear cross-attention block.}
    \label{fig:rwkv_cross}
\end{figure}
The decoder is responsible for generating multi-modal trajectory candidates, where each trajectory is represented as a sequence of $N$ waypoints $\tau = \{ (x_t, y_t, \theta_t) \}_{t = 1}^{N}$. 
Here, $(x_t, y_t)$ and $\theta_t$ denote the position and orientation in the ego-centric coordinate frame, with $N$ representing the prediction horizon.
To decode these trajectories based on the surrounding context, we employ learnable queries to interact with the augmented BEV-ego representations, as shown in Fig.~\ref{fig:whole}(a).
While standard approaches~\cite{yuan2024drama,dao2024transformers,vaswani2017attention} rely on Transformer-based cross-attention to handle sequences of differing lengths, this introduces quadratic complexity, violating LADY’s fully linear design principle and hindering real-time performance with high-resolution inputs.

To resolve this bottleneck, we propose LICA, a generic mechanism that extends linear attention architectures to support cross-modal interaction efficiently, as shown in Fig.~\ref{fig:rwkv_cross}.
Instead of calculating a quadratic attention map, LICA treats cross-attention as a sequence continuation task:
\textbf{Query Encoding:} the learnable queries are first passed through a linear attention block (specially, RWKV-7) to encode their positional dependencies—analogous to self-attention in Transformers—yielding the encoded queries.
\textbf{Sequential Interaction:} we concatenate the context features (BEV-ego representations) and the encoded queries along the sequence dimension to form a unified stream.
\textbf{State-based Conditioning:} this combined sequence is fed into a second RWKV-7 block. 
Crucially, as the recurrent state evolves through the context features, it naturally accumulates semantic information, which is then carried over to condition the subsequent query tokens without additional quadratic cost.
Finally, we extract the last $M$ tokens (corresponding to the queries) as the cross-attention output, as shown in Fig.~\ref{fig:rwkv_cross}.
By reformulating cross-attention as a single-stream linear operation, LICA achieves linear scaling with respect to sequence length, preserving the fully linear attention of the LADY architecture.

\subsection{Diffusion-based Trajectory Decoder}
\begin{figure}[!t]
    \centering
    \includegraphics[width=1.0\linewidth]{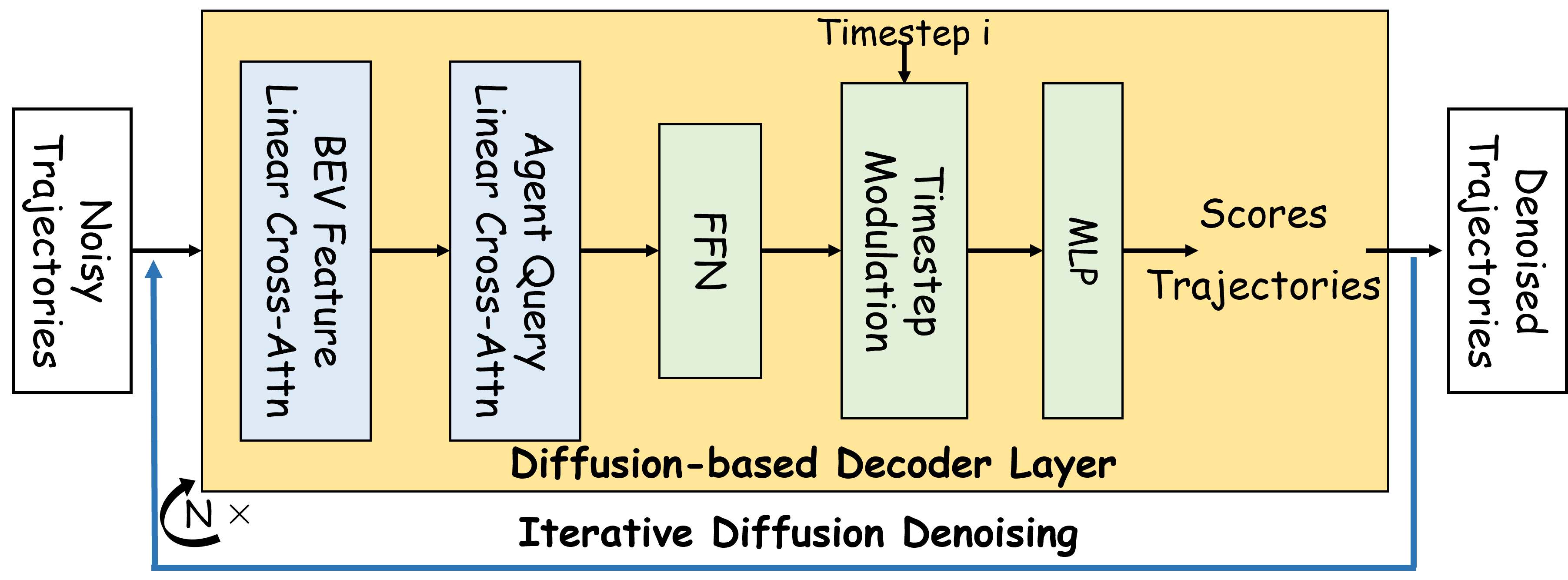}
    \caption{Overall architecture of the diffusion-based decoder layer.}
    \label{fig:diffusion_decoder}
\end{figure}
To capture richer multi-modal trajectory distributions, we implement the truncated diffusion policy adopted from DiffusionDrive~\cite{liao2025diffusiondrive}. 
The diffusion-based decoder takes slightly noisy anchor trajectories as input. 
At each denoising step, it conditions on two key information sources: the fused BEV features and the encoded queries derived from the previous stage, which explicitly capture the context of surrounding agents.
Specifically, we leverage our proposed LICA block to facilitate efficient cross-attention between the noisy anchor trajectories and these context representations (BEV features and agent queries), as illustrated in Fig.~\ref{fig:diffusion_decoder}.
Additionally, inspired by iPad~\cite{guo2025ipad}, the diffusion-based decoder outputs a set of denoised trajectories alongside their confidence scores and two auxiliary heads to enhance planning performance, as shown in Fig.~\ref{fig:whole}.
These scores are distilled from rule-based expert knowledge embedded in NAVSIM v1~\cite{dauner2024navsim}, with the highest-scoring trajectory selected for the final predicted trajectory.
The mapping head estimates on-road and on-route probabilities for multi-modal trajectories, and the prediction head that forecasts the future states related to potential collisions of surrounding agents, enabling the ego-vehicle to perform maneuvers that are both assertive and safe.

%% file: sections/exp.tex
\section{Experiments}
\begin{table*}[!t]
\centering
\caption{Quantitative comparison with state-of-the-art methods on the NAVSIM \texttt{navtest} split. C \& L denote Camera and LiDAR modalities, respectively. ``Multi-frame'' indicates models utilizing temporal sensor inputs for trajectory generation. $T$ represents the sequence length, and $d$ denotes the feature dimension.}
\label{tab:whole}
\begin{threeparttable}
\resizebox{\textwidth}{!}{%
\begin{tabular}{l|cccc|cc|cccccc}
\toprule
\midrule
\multirow{2}{*}{Methods} & \multirow{2}{*}{Modality} & \multirow{2}{*}{Architecture} & \multirow{2}{*}{BackBone} & \multirow{2}{*}{Multi-Frame} & \multicolumn{2}{c|}{Complexity} & \multicolumn{6}{c}{Metrics} \\
& & & & & Time & Memory & NC & DAC & TTC & Comf. & EP & PDMS \\ \midrule
UniAD~\cite{hu2023planning} & Camera & Transformer  & ResNet-34 & \xmark & $\mathcal{O}(T^2 d)$ & $\mathcal{O}(T^2 + Td)$ & 97.8 & 91.9 & 92.9 & 100 & 78.8 & 83.4 \\
Transfuser~\cite{chitta2022transfuser} & C \& L & Transformer  & ResNet-34 & \xmark & $\mathcal{O}(T^2 d)$ & $\mathcal{O}(T^2 + Td)$ & 97.7 & 92.8 & 92.8 & 100 & 79.2 & 84.0 \\
DRAMA~\cite{yuan2024drama}   & C \& L & MT\tnote{b}  & ResNet-34  & \xmark & $\mathcal{O}(T d)$\tnote{c} & $\mathcal{O}(d)$ & 98.0 & 93.1 & 94.8 & 100 & 80.1 & 85.5 \\
VADv2~\cite{chen2024vadv2}   & Camera & Transformer     & ResNet-34   & \xmark & $\mathcal{O}(T^2 d)$  & $\mathcal{O}(T^2 + Td)$ & 97.2 & 89.1 & 91.6 & 100 & 76.1 & 80.9 \\
Hydra-MDP~\cite{li2025hydra}           & C \& L & Transformer     & ResNet-34   & \xmark & $\mathcal{O}(T^2 d)$ & $\mathcal{O}(T^2 + Td)$ & 98.3 & 96.0 & 94.6 & 100 & 78.7 & 86.5 \\
DiffusionDrive~\cite{liao2025diffusiondrive} & C \& L & Transformer & ResNet-34 & \xmark & $\mathcal{O}(T^2 d)$ & $\mathcal{O}(T^2 + Td)$ & 98.2 & 96.2 & 94.7 & 100 & 82.2 & 88.1 \\ 
iPad~\cite{guo2025ipad} & Camera & Transformer & ResNet-34 & \xmark & $\mathcal{O}(T^2 d)$ & $\mathcal{O}(T^2 + Td)$ & \textbf{98.6} & \textbf{98.3} & \textbf{94.9} & 100 & 88.0 & \textbf{91.7} \\ 
\midrule
DiffusionDrive\tnote{a}& C \& L & Transformer & ResNet-34 & \cmark & $\mathcal{O}(T^2 d)$ & $\mathcal{O}(T^2 + Td)$ & 97.5 & 97.1 & 93.2 & 100 & 86.1 & 89.3 \\ 
\textbf{LADY (Ours)} & C \& L & Linear Attention & ResNet-34 & \cmark & $\mathcal{O}(T d)$ & $\mathcal{O}(d)$ & 98.0 & 97.3 & 94.0 & \textbf{100} & \textbf{88.6} & 90.9  \\ 
\bottomrule
\end{tabular}
}
\begin{tablenotes}
    \footnotesize
    \item[a] Multi-frame version  \; $^\text{b}$ Mamba \& Transformer
    \item[c] DRAMA’s Transformer-based cross-attention makes its time and memory complexity slightly higher than the table values and LADY
\end{tablenotes}
\end{threeparttable}
\end{table*}

\subsection{Experimental Settings}
\noindent\textbf{Experimental Benchmarks}
To ensure a comprehensive assessment, we evaluate LADY on two complementary benchmarks: \textbf{NAVSIM}~\cite{dauner2024navsim} for open-loop planning accuracy and \textbf{Bench2Drive}~\cite{jia2024bench2drive} for closed-loop driving capability.
Consistent with standard works, we report the Predictive Driver Model Score (PDMS) for NAVSIM and the multi-ability results for Bench2Drive, enabling a holistic view of both planning quality and execution robustness.

\begin{table}
\centering
\caption{Upper-bound performance analysis.}
\label{tab:limit}
\resizebox{\columnwidth}{!}{%
\begin{tabular}{lcccccc}
\toprule
\midrule
Methods & NC & DAC & TTC & Comf. & EP  & PDMS\\ \midrule
LADY  & 98.0 & 97.3 & 94.0 & 100  & 88.6 & 90.9 \\
LADY (Best-of-N)  & 100 & 100 & 100 & 100  & 98.13 & 99.2 \\ \midrule
Human  & 100 & 100 & 100 & 100  & 87.5 & 94.8 \\
\bottomrule
\end{tabular}
}
\end{table}

\noindent\textbf{Implementation Details}
LADY takes multi-frame camera and LiDAR features, along with the ego-vehicle's state and noisy anchor trajectories, as inputs.
Specifically, LADY is trained with a fixed historical context of $10$ frames (spanning $5.0$ seconds at $2\text{ Hz}$); 
sequences shorter than this horizon are prepended with zero tensors to ensure uniform temporal length.
The final output consists of an $8$-waypoint trajectory spanning $4$ seconds (sampled at $2\text{ Hz}$), where each waypoint comprises 2D coordinates $(x, y)$ and heading $\theta$.
The model is trained from scratch for $100$ epochs on the standard training splits of the respective benchmarks (i.e., \texttt{navtrain} for NAVSIM and the official \texttt{Base} training set for Bench2Drive).
We employ the Adam optimizer with a total batch size of $4$, distributed across $4$ NVIDIA RTX 4090 GPUs ($48$ GB each). 
The learning rate is initialized at $1 \times 10^{-4}$ and managed by a \texttt{ReduceLROnPlateau} scheduler ($\texttt{patience} = 10$, $\texttt{factor} = 0.5$).
For the diffusion-based trajectory decoder, we employ a cascaded architecture with $2$ diffusion decoder layers and adopt the truncated diffusion policy from DiffusionDrive~\cite{liao2025diffusiondrive}.
During training, we truncate the diffusion schedule to $50$ out of $1000$ steps to diffuse $100$ clustered anchor trajectories, thereby enriching multi-modal representations. 
Notably, to ensure real-time performance, we perform only $2$ denoising steps during inference and select the highest-scoring trajectory for evaluation.
To validate deployment feasibility, all efficiency metrics are measured on an NVIDIA Jetson AGX Orin ($32$ GB), a standard embedded platform for autonomous systems.
Besides, our RWKV implementation incorporates the optimized Flash Linear Attention kernels~\cite{yang2024fla}.

\begin{table*}
\centering
\caption{Closed-loop and multi-ability results for varying end-to-end methods on the Bench2Drive benchmark.}
\label{tab:bench2drive_results}
\begin{threeparttable}
\resizebox{\textwidth}{!}{%
\begin{tabular}{l|cccc|cccccc}
\toprule
\midrule
\multirow{2}{*}{Method} & \multicolumn{4}{c|}{Closed-loop Results} & \multicolumn{6}{c}{Multi-Ability Results (\%)} \\
& Comfortness & Efficiency & Success Rate (\%) & Driving Score\(\uparrow\) & Merging & Overtaking\(\uparrow\) & Emergency Brake
& Give Way\(\uparrow\) & Traffic Sign & Mean\(\uparrow\) \\ \midrule
UniAD-Tiny~\cite{hu2023planning} & 47.04 & 123.92 & 13.18 & 40.73 & 8.89 & 9.33 & 20.00 & 20.00 & 15.43 & 14.73 \\
UniAD-Base~\cite{hu2023planning} & 43.58 & 129.21 & 16.36 & 45.81 & 14.10 & 17.78 & 21.67 & 10.00 & 14.21 & 15.55 \\
VAD~\cite{jiang2023vad} & 46.01 & 157.94& 15.00 & 42.35 & 8.11 & 24.44 & 18.64 & 20.00 & 19.15 & 18.07 \\
DriveTransformer~\cite{jia2025drivetransformer} & 20.78 & 100.64 & 35.01 & 63.46 & 17.57 & 35.00 & 48.36 & 40.00 & 52.10 & 38.60 \\
iPad~\cite{guo2025ipad} & 28.21 & \textbf{161.31} & \textbf{35.91} & 65.02 & \textbf{30.00} & 20.00 & \textbf{53.33}
& 60.00 & 49.47 & 42.56 \\
\textbf{LADY(1 frame)} & 17.55 & 141.31 & 14.55 & 50.06 & 20.00 & 4.44 & 8.33 & 50.00 & 24.21 & 21.40\\
\textbf{LADY(10 frames)} & 36.26 & 155.39 & 25.91 & 57.52 & 26.25 & 33.33 & 21.67 & 50.00 & 28.42 & 31.93 \\
\textbf{LADY(Infinite frames)} & 12.13 & 155.57 & 33.02 & \textbf{65.12} & 24.36 & \textbf{55.81} & 35.71 & \textbf{70.00} & 28.80 & \textbf{42.94} \\
\midrule
TCP*~\cite{wu2022trajectory} & 47.80 & 54.26 & 15.00 & 40.70 & 16.18 & 20.00 & 20.00 & 10.00 & 6.99 & 14.63 \\
TCP-ctrl* & \textbf{51.51}  & 55.97 & 7.27 & 30.47 & 10.29 & 4.44 & 10.00 & 10.00 & 6.45 & 8.23 \\
TCP-traj* & 18.08  & 76.54 & 30.00 & 59.90 & 8.89 & 24.29 & 51.67 & 40.00 & 46.28 & 34.23 \\
TCP-traj w/o distill. & 22.96 & 78.78 & 20.45 & 49.30 & 13.53 & 16.67 & 40.00 & 50.00 & 28.72 & 28.51 \\
ThinkTwice*~\cite{jia2023think} & 16.22 & 69.33 & 31.23 & 62.44 & 27.38 & 18.42 & 35.82 & 50.00 & 54.23 & 37.17 \\
DriveAdapter*~\cite{jia2023driveadapter} & 16.01 & 70.22 & 33.08 & 64.22 & 28.82 & 26.38 & 48.76 & 50.00 & \textbf{56.43} & 42.08 \\
\bottomrule
\end{tabular}
}
\begin{tablenotes}
    \footnotesize
    \item[*] denotes expert feature distillation
\end{tablenotes}
\end{threeparttable}
\end{table*}

\subsection{Comparison with State-of-the-Art}
\begin{table}
\centering
\caption{Performance comparison of inference time and memory usage across different frame inputs ($N$).}
\label{tab:performance_comparison}
\begin{threeparttable}
\resizebox{\linewidth}{!}{%
\begin{tabular}{lcccccc}
\toprule
\midrule
\multirow{2}{*}{\textbf{Model}} & \multicolumn{3}{c}{\textbf{Inference Time (ms)} $\downarrow$} & \multicolumn{3}{c}{\textbf{Memory Usage (MB)} $\downarrow$} \\ 
\cmidrule(lr){2-4} \cmidrule(lr){5-7}
 & \textbf{$N=10$} & \textbf{$N=30$} & \textbf{$N=50$} & \textbf{$N=10$} & \textbf{$N=30$} & \textbf{$N=50$} \\ 
\midrule
Transfuser (Multi-Frame) & 419 & 1816 & 5729 & 673 & 3444 & 8711 \\
DiffusionDrive (Multi-Frame) & 473 & 1779 & 5608 & 690 & 3460 & 8729 \\
DRAMA (Multi-Frame) & 413 & 784 & 1144 & 610 & 1537 & 2428 \\
\midrule
LADY (Variant)\tnote{*} & 481 & 910 & 1330 & 467 & 917 & 1364 \\ 
\textbf{LADY (Ours)} & \textbf{187} & \textbf{187} & \textbf{187} & \textbf{350} & \textbf{410} & \textbf{480} \\ 
\bottomrule
\end{tabular}%
}
\begin{tablenotes}
    \footnotesize
    \item[*] uses Transformer-based standard cross-attention
\end{tablenotes}
\end{threeparttable}
\end{table}
Table~\ref{tab:whole} compares LADY with SOTA methods on the NAVSIM \texttt{navtest} split for open-loop evaluations. 
All models use the standard ResNet-34 backbone, LADY achieves a competitive overall PDMS of $90.9$. 
By effectively integrating multi-frame sensor information via the diffusion-based decoder, our method outperforms most existing baselines. 
Notably, compared to approaches like Transfuser and DRAMA, the significant improvement in the EP metric highlights LADY's ability to plan more efficient and human-like maneuvers.

Despite the gap with iPad~\cite{guo2025ipad}, our ``Best-of-N'' evaluation yields a remarkable PDMS of $99.2$, where the optimal trajectory is selected based on the NAVSIM ground-truth evaluator rather than the predicted scores, as shown in Table~\ref{tab:limit}.
This confirms that our backbone generates superior candidates, currently limited only by the scoring head's ranking ability rather than generative capacity. 
Furthermore, we attribute the modest open-loop gains to NAVSIM's lack of samples requiring long-range temporal context—the core strength of our linear attention design. 
Consequently, to rigorously validate LADY in dynamic environments where long-term dependency is critical, we proceed to the closed-loop evaluation on Bench2Drive.

To systematically analyze the impact of temporal context length on driving performance, we instantiate LADY in three variants with distinct temporal feature fusion strategies:
\textbf{Single-frame (1 frame):} processes only the camera and LiDAR features from the current timestamp, ignoring all historical context;
\textbf{Short-term (10 frames):} fuses features from the current frame and the preceding $9$ frames, providing a limited temporal receptive field;
\textbf{Long-term (Infinite frames):} maintains a continuous temporal hidden state that aggregates feature information from the entire history sequence.
This variant fully leverages the RNN-like property of LADY to model unbounded temporal dependencies.
As detailed in Table~\ref{tab:bench2drive_results}, LADY significantly benefits from long-term context integration, outperforming baseline methods. 
Crucially, thanks to the linear attention mechanism, this performance gain is achieved with constant inference complexity, introducing no additional computational overhead compared to the short-term variants.

\subsection{Efficiency Analysis}
\begin{figure}
    \centering
    \begin{subfigure}{0.49\linewidth}
        \centering
        \includegraphics[width=0.98\linewidth]{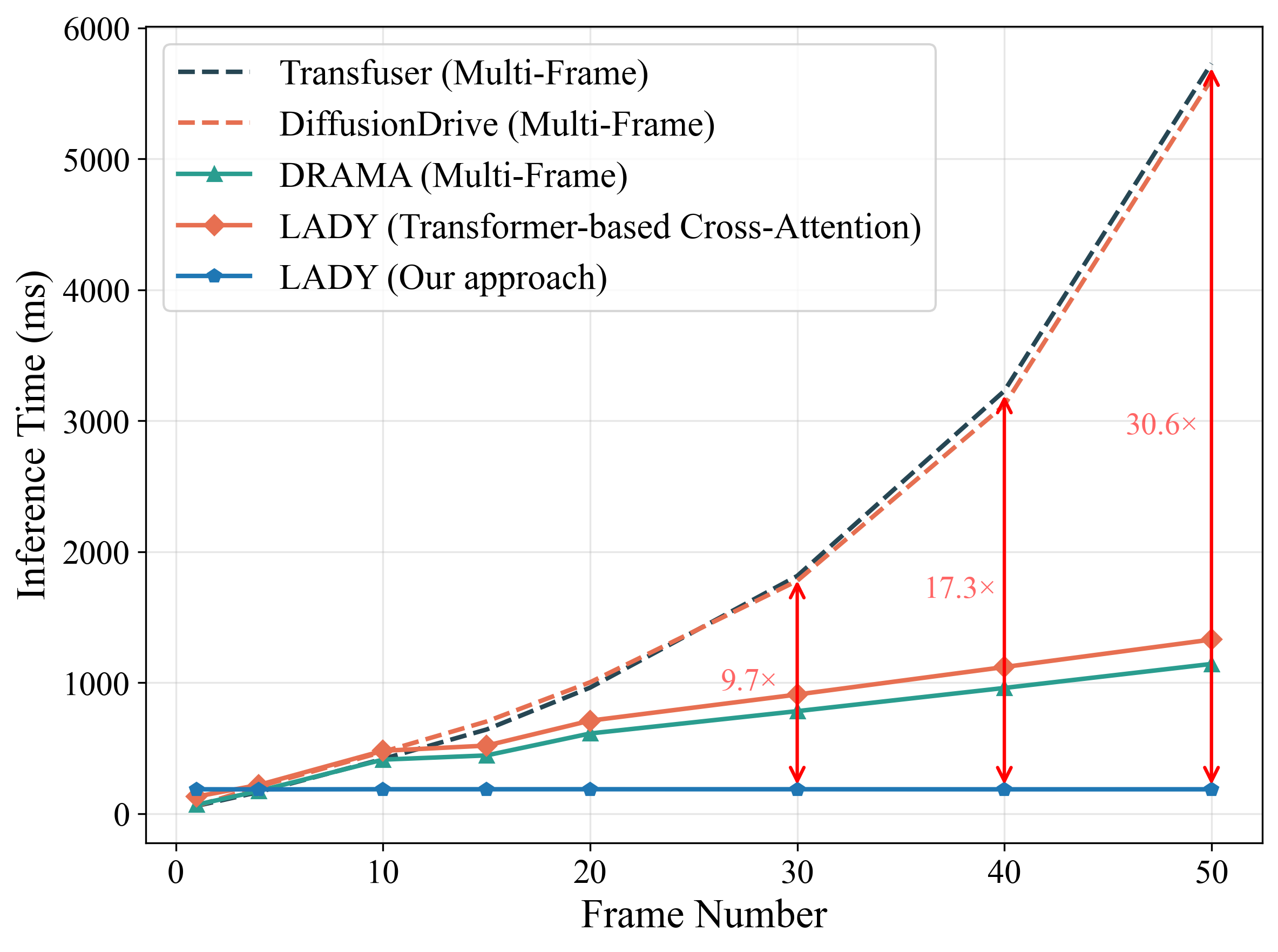}
        \label{subfig:time}
    \end{subfigure}
    \centering
    \begin{subfigure}{0.49\linewidth}
        \centering
        \includegraphics[width=0.98\linewidth]{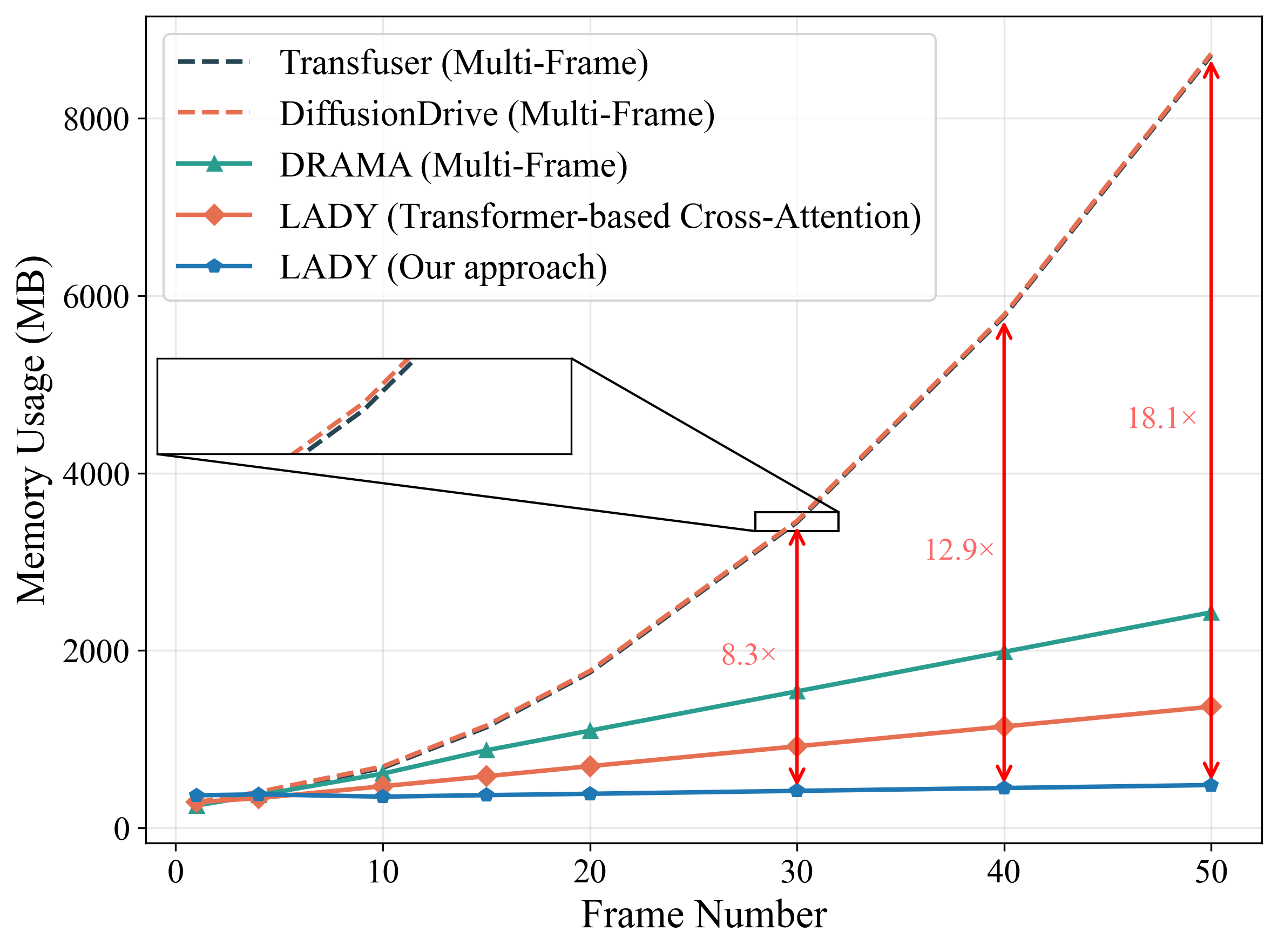}
        \label{subfig:memory}
    \end{subfigure}
    \caption{Computational efficiency analysis. Comparison of inference latency and memory usage between LADY and baseline methods.}
    \label{fig:cmp}
\end{figure}

Fig.~\ref{fig:cmp} and Table~\ref{tab:performance_comparison} present a comparative analysis of inference time and memory usage for LADY, Transfuser, DRAMA, and DiffusionDrive as the input frame count increases. 
All measurements are conducted on the NVIDIA Orin platform to simulate real-world edge deployment.
During inference, LADY leverages its recurrent formulation to operate in a sequential mode, updating a fixed-size compact temporal hidden state frame-by-frame. 
Consequently, its computational graph remains uncoupled from the accumulated history length, resulting in constant ($O(1)$) latency and memory footprint regardless of the input frame count, consistently maintaining approximately $187$ ms inference time and $400$ MB memory usage.
In contrast, Transformer-based architectures (Transfuser and DiffusionDrive) require recomputing attention over the entire history of tokens, causing inference time and memory consumption to grow sharply with each additional frame.
Similarly, DRAMA—and the LADY variant using standard cross-attention (w/o LICA)—fail to achieve this efficiency; due to their lack of a fully linear architecture, and their computational overhead also escalates as the input sequence lengthens.
This analysis confirms that LADY's constant-time, constant-memory profile is uniquely suited for real-time end-to-end autonomous driving, where fusing long-range temporal context is critical.

\subsection{Ablation Studies}
\begin{table}
\centering
\caption{Ablation study on temporal generalization. Performance of the model trained with a fixed history window when evaluating on varying inference sequence lengths.}
\label{tab:ablation}
\begin{tabular}{ccccccc}
\toprule
\midrule
Frame Num. & NC\(\uparrow\)  & DAC\(\uparrow\)  & TTC\(\uparrow\)  & Comf. & EP\(\uparrow\)   & PDMS\(\uparrow\) \\ \midrule
1  & 97.4 & 95.5 & 91.8 & 100  & 83.2 & 86.8 \\
4  & 97.6 & 96.9 & 92.9 & 100  & 86.8 & 89.5 \\
6  & 97.7 & 96.9 & 92.9 & 100  & 87.8 & 90.0 \\
8  & 97.7 & 97.1 & 93.3 & 100  & 88.2 & 90.3 \\
10 & 97.9 & 97.1 & 93.6 & 100  & 88.5 & 90.6 \\
12 & 97.9 & 97.2 & 93.7 & 100  & 88.5 & 90.7 \\
15 & \textbf{98.0} & \textbf{97.3} & 94.0 & \textbf{100} & \textbf{88.6} & \textbf{90.9} \\
20 & 98.0 & 97.2 & \textbf{94.1} & 100 & 88.5 & \textbf{90.9} \\ 
\bottomrule
\end{tabular}%
\end{table}

\begin{figure}[!t]
    \centering
    \includegraphics[width=0.9\linewidth]{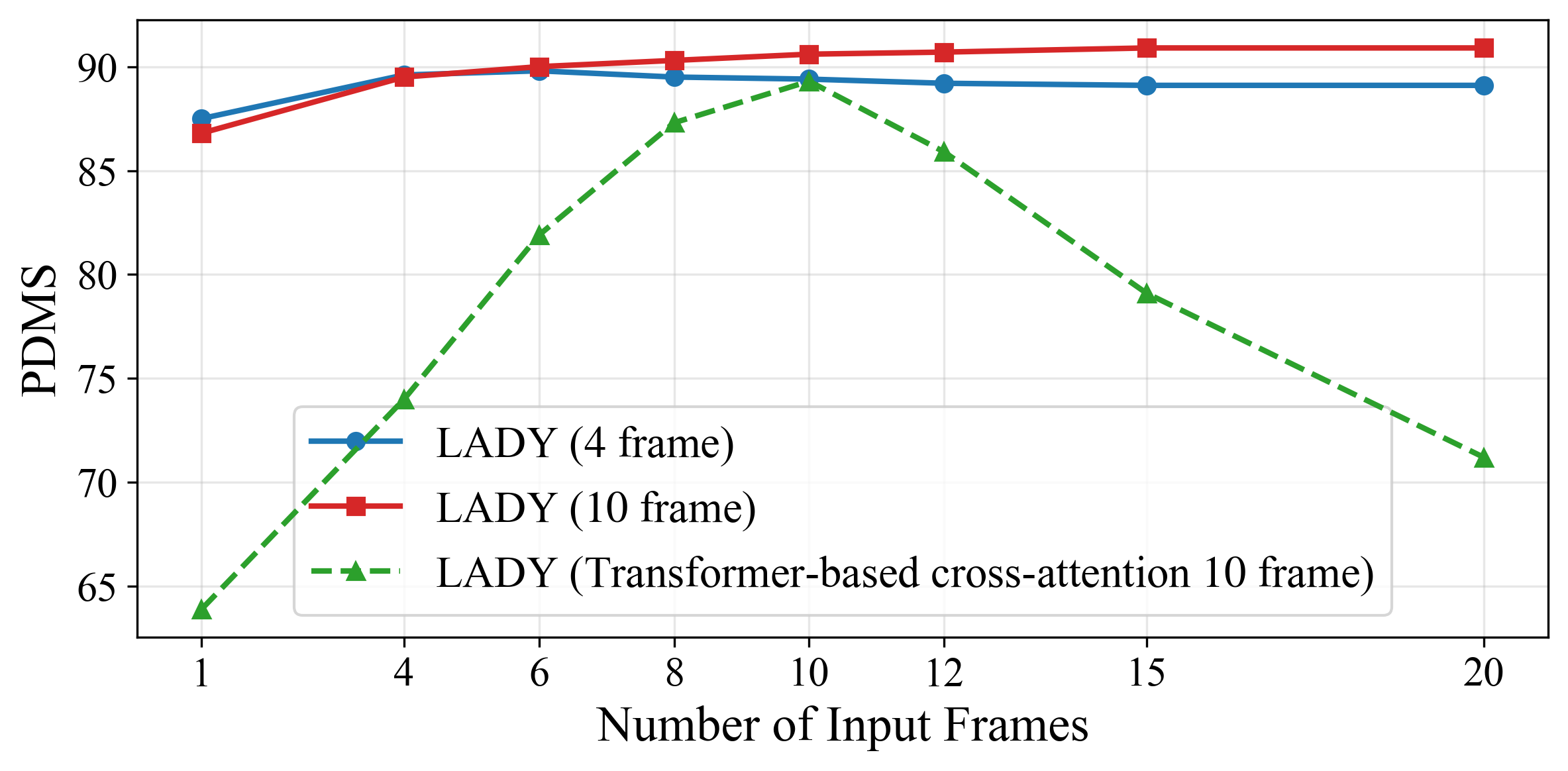}
    \caption{Ablation study on temporal context length and model architecture.}
    \label{fig:ablation_frames}
\end{figure}
\noindent\textbf{Effectiveness of Historical Frame Count}
First, to validate the fundamental benefit of temporal integration, we extend the baseline DiffusionDrive to a multi-frame setting. 
This results in a PDMS improvement from $88.1$ to $89.3$, confirming the effectiveness of incorporating historical information.
Subsequently, we investigate LADY's temporal generalization capability to further validate the effectiveness of temporal information.
We trained LADY with a fixed history window of $10$ frames but evaluated it using varying input lengths during inference, as detailed in Table~\ref{tab:ablation}. 
For sequences shorter than the training window, zero-padding is applied to the front.
The results indicate a performance drop when input frames are insufficient ($<10$), as expected. 
However, a crucial advantage of our linear attention design emerges when inference sequences exceed the training length. 
LADY seamlessly updates its temporal hidden state to incorporate the additional history, yielding performance that even surpasses the baseline trained on fixed frames. 
This demonstrates LADY's ability to perform temporal extrapolation, making it highly suitable for continuous, real-time driving scenarios.

\begin{figure*}
    \centering
    \begin{minipage}[b]{0.64\linewidth} 
        \centering
        \begin{subfigure}{\linewidth}
            \centering
            \includegraphics[width=\linewidth]{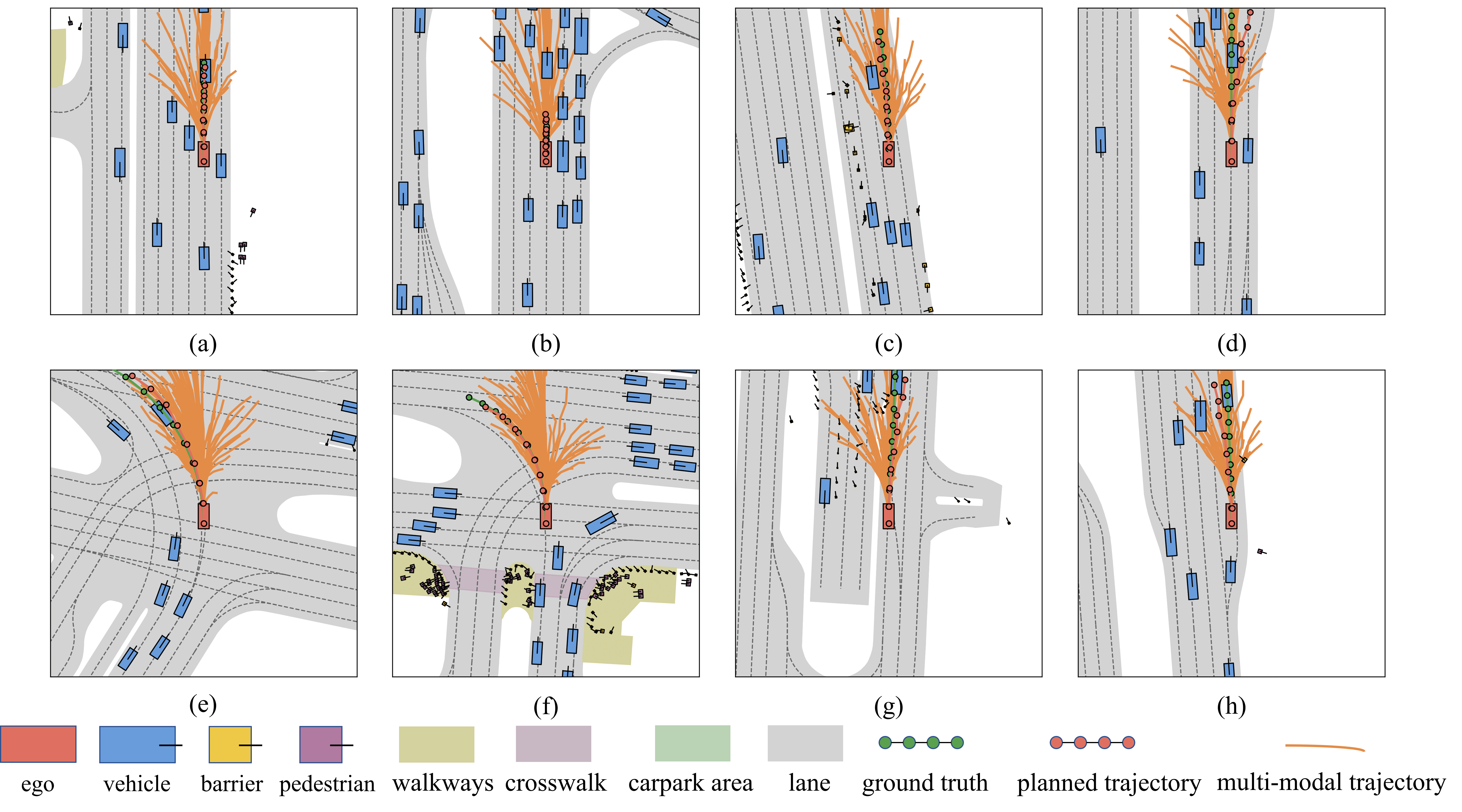}
            \caption{Representative scenarios on NAVSIM.}
            \label{subfig:navsim_fig}
        \end{subfigure}
    \end{minipage}
    \hfill
    \begin{minipage}[b]{0.34\linewidth}
        \centering
        \begin{subfigure}{\linewidth}
            \centering
            \includegraphics[width=\linewidth]{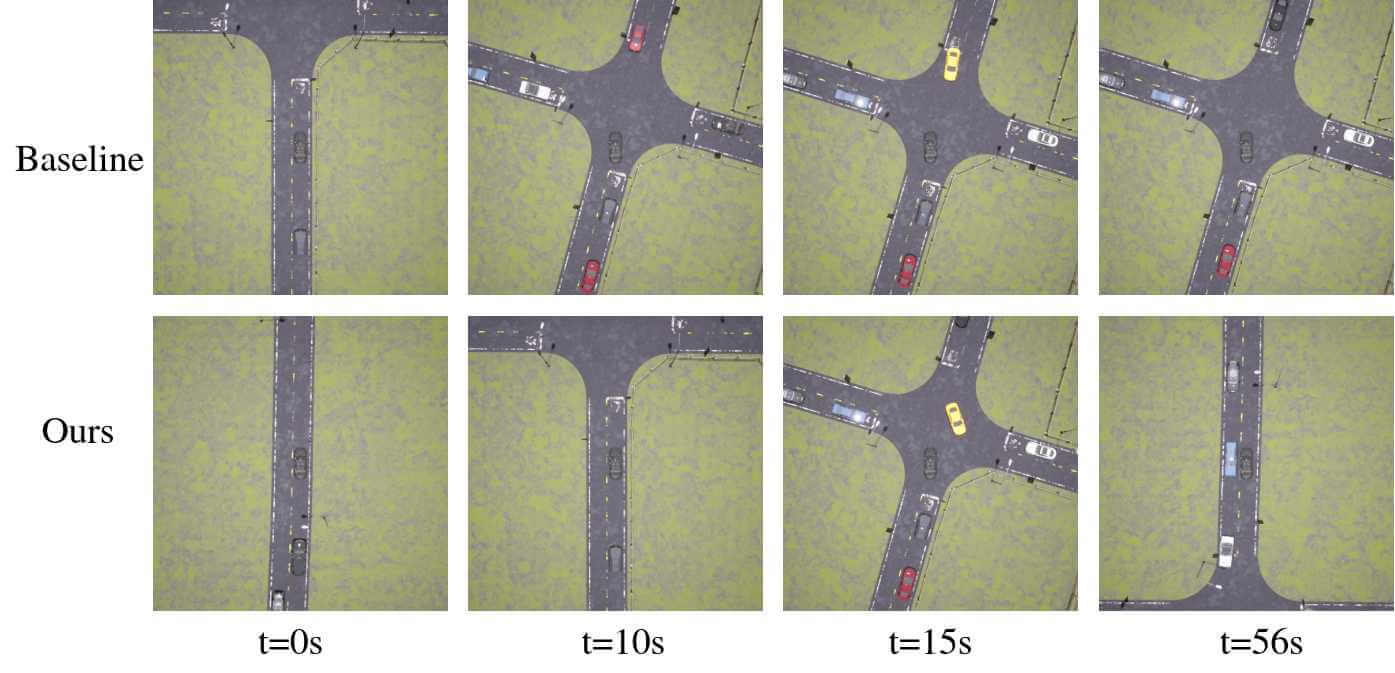}
            \caption{Intersection scenario (Bench2Drive).}
            \label{subfig:intersection}
        \end{subfigure}
        
        \vspace{10pt}
        \begin{subfigure}{\linewidth}
            \centering
            \includegraphics[width=\linewidth]{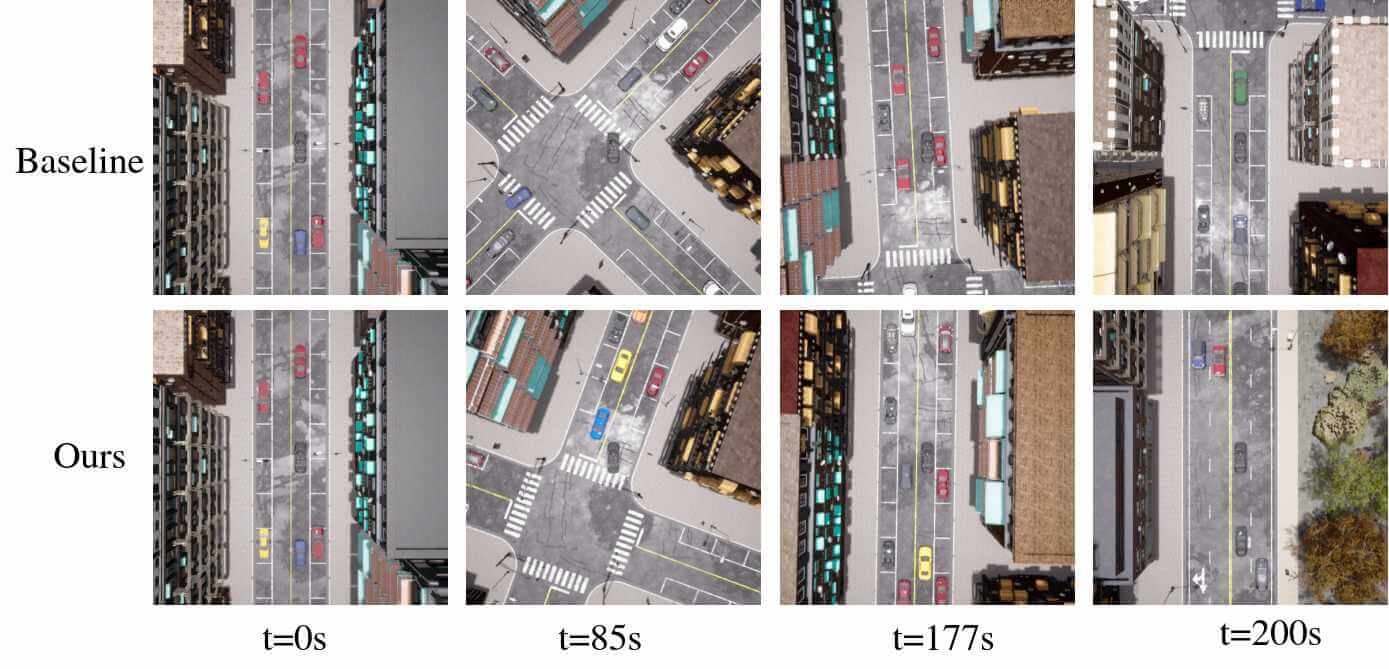} 
            \caption{Long-horizon route scenario (Bench2Drive).}
            \label{subfig:longtime}
        \end{subfigure}
    \end{minipage}
    \caption{\textbf{Qualitative analysis across open-loop and closed-loop benchmarks.} 
    (a) Visualization of representative planning results of LADY on the \textbf{NAVSIM} dataset. 
    (b)-(c) Comparative evaluation between LADY and the SOTA method iPad~\cite{guo2025ipad} on \textbf{Bench2Drive}, focusing on (b) complex intersection handling and (c) long-horizon route tracking. LADY demonstrates superior capability in handling dynamic agents and maintaining long-term consistency.}
    \label{fig:representative_scenario}
\end{figure*}
Additionally, we analyze the impact of the training horizon on model performance.
We compare a model trained with a $4$-frame history to our $10$-frame baseline.
As shown in Fig.~\ref{fig:ablation_frames}, training with a richer historical context yields significant performance gains. 
More importantly, a longer training horizon enhances the model's capability for temporal extrapolation, allowing it to maintain robustness even when the inference sequence length exceeds the training setting.

\noindent\textbf{Effectiveness of the LICA Module}
Finally, we validate the efficacy of our proposed LICA module. 
For a fair comparison, we replaced the LICA block in the decoder with a standard Transformer-based cross-attention mechanism, training both models with the same $10$-frame history.
As illustrated in Fig.~\ref{fig:ablation_frames}, the LICA-based implementation consistently outperforms the standard cross-attention baseline across all test settings. 
More importantly, LICA exhibits superior temporal extrapolation capabilities that are absent in the standard approach. 
When the inference sequence length exceeds the training horizon ($>10$ frames), LICA's performance remains stable or even improves. 
In stark contrast, the standard cross-attention model suffers a sharp performance degradation in these scenarios, failing to generalize beyond the fixed training window.
Consequently, LADY empowers the efficient aggregation of long-range temporal context from multi-frame camera and LiDAR features—a capability that remains unattainable for standard Transformer-based approaches.

\subsection{Qualitative Analysis}
To qualitatively validate LADY, we visualize representative driving scenarios from both open-loop and closed-loop benchmarks in Fig.~\ref{fig:representative_scenario}.
\textbf{Open-loop Performance (NAVSIM):} 
As shown in Fig.~\ref{fig:representative_scenario}(\subref{subfig:navsim_fig}), LADY generates multi-modal, context-aware trajectories across various scenarios, such as car-following, overtaking, and turning.
By selecting the optimal trajectory based on confidence scores, our model demonstrates safe yet assertive behaviors, effectively avoiding the excessive conservatism often observed in baseline methods.
\textbf{Closed-loop Performance (Bench2Drive):}
Figs.~\ref{fig:representative_scenario}(\subref{subfig:intersection}) and (\subref{subfig:longtime}) compare LADY with the SOTA method iPad~\cite{guo2025ipad}.
In the complex intersection scenario (\subref{subfig:intersection}), iPad exhibits prolonged stationary behavior due to overly conservative planning based on limited context.
Similarly, in the long-horizon route task (\subref{subfig:longtime}), iPad suffers from hesitation at junctions, leading to mission failure.
In contrast, LADY successfully completes both missions. 
By leveraging the accumulated historical observation sequences, our model infers richer interaction contexts, enabling smoother and more consistent decision-making in dynamic environments.

%% file: sections/conclusion.tex
\section{Conclusion}
In this work, we proposed LADY, the first fully linear attention-based model designed for end-to-end autonomous driving. 
To ensure full linearity throughout the architecture, we introduced the lightweight LICA mechanism for linear cross-attention.
By leveraging these designs, LADY effectively fuses long-range camera and LiDAR features with constant computational and memory overhead, overcoming the scaling limitations of traditional Transformers. 
Extensive experiments on the NAVSIM and Bench2Drive benchmarks demonstrate that LADY achieves significantly lower inference costs on long sequences while delivering superior planning performance.
Future work will focus on designing a more sophisticated scoring module to fully unleash the model's generative capabilities. Additionally, we plan to explore alternative linear attention architectures and validate LADY's robustness in real-world driving scenarios.

%% file: ref.bib
@incollection{widrow1988adaptive,
  title={Adaptive switching circuits},
  author={Widrow, Bernard and Hoff, Marcian E},
  booktitle={Neurocomputing: foundations of research},
  pages={123--134},
  publisher = {MIT Press},
  year={1988}
}

@inproceedings{he2016deep,
 author = {He, Kaiming and Zhang, Xiangyu and Ren, Shaoqing and Sun, Jian},
 title = {Deep Residual Learning for Image Recognition},
 booktitle = {Proceedings of the IEEE Conference on Computer Vision and Pattern Recognition (CVPR)},
 year = {2016}
}

@article{vaswani2017attention,
  title={Attention is all you need},
  author={Vaswani, Ashish and Shazeer, Noam and Parmar, Niki and Uszkoreit, Jakob and Jones, Llion and Gomez, Aidan N and Kaiser, {\L}ukasz and Polosukhin, Illia},
  journal={Advances in neural information processing systems},
  volume={30},
  year={2017}
}

@inproceedings{schlag2021linear,
  title={Linear transformers are secretly fast weight programmers},
  author={Schlag, Imanol and Irie, Kazuki and Schmidhuber, J{\"u}rgen},
  booktitle={International conference on machine learning},
  pages={9355--9366},
  year={2021}
}

@article{wu2022trajectory,
  title={Trajectory-guided control prediction for end-to-end autonomous driving: A simple yet strong baseline},
  author={Wu, Penghao and Jia, Xiaosong and Chen, Li and Yan, Junchi and Li, Hongyang and Qiao, Yu},
  journal={Advances in Neural Information Processing Systems},
  volume={35},
  pages={6119--6132},
  year={2022}
}

@article{chitta2022transfuser,
  title={Transfuser: Imitation with transformer-based sensor fusion for autonomous driving},
  author={Chitta, Kashyap and Prakash, Aditya and Jaeger, Bernhard and Yu, Zehao and Renz, Katrin and Geiger, Andreas},
  journal={IEEE transactions on pattern analysis and machine intelligence},
  volume={45},
  number={11},
  pages={12878--12895},
  year={2022}
}

@article{lu2022real,
  author={Lu, Yongqiang and Ma, Hongjie and Smart, Edward and Yu, Hui},
  journal={IEEE Transactions on Intelligent Transportation Systems}, 
  title={Real-Time Performance-Focused Localization Techniques for Autonomous Vehicle: A Review}, 
  year={2022},
  volume={23},
  number={7},
  pages={6082-6100}
}

@inproceedings{jia2023driveadapter,
  title={Driveadapter: Breaking the coupling barrier of perception and planning in end-to-end autonomous driving},
  author={Jia, Xiaosong and Gao, Yulu and Chen, Li and Yan, Junchi and Liu, Patrick Langechuan and Li, Hongyang},
  booktitle={Proceedings of the IEEE/CVF International Conference on Computer Vision},
  pages={7953--7963},
  year={2023}
}

@inproceedings{jia2023think,
  title={Think twice before driving: Towards scalable decoders for end-to-end autonomous driving},
  author={Jia, Xiaosong and Wu, Penghao and Chen, Li and Xie, Jiangwei and He, Conghui and Yan, Junchi and Li, Hongyang},
  booktitle={Proceedings of the IEEE/CVF Conference on Computer Vision and Pattern Recognition},
  pages={21983--21994},
  year={2023}
}

@inproceedings{hu2023planning,
  title={Planning-oriented autonomous driving},
  author={Hu, Yihan and Yang, Jiazhi and Chen, Li and Li, Keyu and Sima, Chonghao and Zhu, Xizhou and Chai, Siqi and Du, Senyao and Lin, Tianwei and Wang, Wenhai and others},
  booktitle={Proceedings of the IEEE/CVF conference on computer vision and pattern recognition},
  pages={17853--17862},
  year={2023}
}

@article{chi2023diffusion,
  title={Diffusion policy: Visuomotor policy learning via action diffusion},
  author={Chi, Cheng and Xu, Zhenjia and Feng, Siyuan and Cousineau, Eric and Du, Yilun and Burchfiel, Benjamin and Tedrake, Russ and Song, Shuran},
  journal={The International Journal of Robotics Research},
  pages={02783649241273668},
  year={2023}
}

@article{ye2023fusionad,
  title={Fusionad: Multi-modality fusion for prediction and planning tasks of autonomous driving},
  author={Ye, Tengju and Jing, Wei and Hu, Chunyong and Huang, Shikun and Gao, Lingping and Li, Fangzhen and Wang, Jingke and Guo, Ke and Xiao, Wencong and Mao, Weibo and others},
  journal={arXiv preprint arXiv:2308.01006},
  year={2023}
}

@inproceedings{jiang2023vad,
  title={Vad: Vectorized scene representation for efficient autonomous driving},
  author={Jiang, Bo and Chen, Shaoyu and Xu, Qing and Liao, Bencheng and Chen, Jiajie and Zhou, Helong and Zhang, Qian and Liu, Wenyu and Huang, Chang and Wang, Xinggang},
  booktitle={Proceedings of the IEEE/CVF International Conference on Computer Vision},
  pages={8340--8350},
  year={2023}
}

@article{teng2023motion,
  author={Teng, Siyu and Hu, Xuemin and Deng, Peng and Li, Bai and Li, Yuchen and Ai, Yunfeng and Yang, Dongsheng and Li, Lingxi and Xuanyuan, Zhe and Zhu, Fenghua and Chen, Long},
  journal={IEEE Transactions on Intelligent Vehicles}, 
  title={Motion Planning for Autonomous Driving: The State of the Art and Future Perspectives}, 
  year={2023},
  volume={8},
  number={6},
  pages={3692-3711}
}

@article{chib2023recent,
  title={Recent advancements in end-to-end autonomous driving using deep learning: A survey},
  author={Chib, Pranav Singh and Singh, Pravendra},
  journal={IEEE Transactions on Intelligent Vehicles},
  volume={9},
  number={1},
  pages={103--118},
  year={2023},
  publisher={IEEE}
}

@inproceedings{wang2023unitr,
  author = {Wang, Haiyang and Tang, Hao and Shi, Shaoshuai and Li, Aoxue and Li, Zhenguo and Schiele, Bernt and Wang, Liwei},
  booktitle = {2023 IEEE/CVF International Conference on Computer Vision (ICCV)},
  title = {UniTR: A Unified and Efficient Multi-Modal Transformer for Bird’s-Eye-View Representation},
  year = {2023},
  pages = {6769-6779},
  publisher = {IEEE Computer Society},
}

@article{gu2023mamba,
  title={Mamba: Linear-time sequence modeling with selective state spaces},
  author={Gu, Albert and Dao, Tri},
  journal={arXiv preprint arXiv:2312.00752},
  year={2023}
}

@article{peng2023rwkv,
  title={Rwkv: Reinventing rnns for the transformer era},
  author={Peng, Bo and Alcaide, Eric and Anthony, Quentin and Albalak, Alon and Arcadinho, Samuel and Biderman, Stella and Cao, Huanqi and Cheng, Xin and Chung, Michael and Grella, Matteo and others},
  journal={arXiv preprint arXiv:2305.13048},
  year={2023}
}

@inproceedings{dauner2024navsim,
 author = {Dauner, Daniel and Hallgarten, Marcel and Li, Tianyu and Weng, Xinshuo and Huang, Zhiyu and Yang, Zetong and Li, Hongyang and Gilitschenski, Igor and Ivanovic, Boris and Pavone, Marco and Geiger, Andreas and Chitta, Kashyap},
 booktitle = {Advances in Neural Information Processing Systems},
 pages = {28706--28719},
 title = {NAVSIM: Data-Driven Non-Reactive Autonomous Vehicle Simulation and Benchmarking},
 volume = {37},
 year = {2024}
}

@article{li2024hydra,
  title={Hydra-mdp: End-to-end multimodal planning with multi-target hydra-distillation},
  author={Li, Zhenxin and Li, Kailin and Wang, Shihao and Lan, Shiyi and Yu, Zhiding and Ji, Yishen and Li, Zhiqi and Zhu, Ziyue and Kautz, Jan and Wu, Zuxuan and others},
  journal={arXiv preprint arXiv:2406.06978},
  year={2024}
}

@article{chen2024vadv2,
  title={Vadv2: End-to-end vectorized autonomous driving via probabilistic planning},
  author={Chen, Shaoyu and Jiang, Bo and Gao, Hao and Liao, Bencheng and Xu, Qing and Zhang, Qian and Huang, Chang and Liu, Wenyu and Wang, Xinggang},
  journal={arXiv preprint arXiv:2402.13243},
  year={2024}
}

@article{chen2026video,
  title={Video mamba suite: State space model as a versatile alternative for video understanding},
  author={Chen, Guo and Huang, Yifei and Xu, Jilan and Pei, Baoqi and Wang, Jiahao and Chen, Zhe and Li, Zhiqi and Lu, Tong and Wang, Limin},
  journal={International Journal of Computer Vision},
  volume={134},
  number={1},
  pages={20},
  year={2026},
  publisher={Springer}
}

@inproceedings{liao2025diffusiondrive,
  title={Diffusiondrive: Truncated diffusion model for end-to-end autonomous driving},
  author={Liao, Bencheng and Chen, Shaoyu and Yin, Haoran and Jiang, Bo and Wang, Cheng and Yan, Sixu and Zhang, Xinbang and Li, Xiangyu and Zhang, Ying and Zhang, Qian and others},
  booktitle={Proceedings of the Computer Vision and Pattern Recognition Conference},
  pages={12037--12047},
  year={2025}
}

@article{dao2024transformers,
  title={Transformers are ssms: Generalized models and efficient algorithms through structured state space duality},
  author={Dao, Tri and Gu, Albert},
  journal={arXiv preprint arXiv:2405.21060},
  year={2024}
}

@article{peng2024eagle,
  title={Eagle and finch: Rwkv with matrix-valued states and dynamic recurrence},
  author={Peng, Bo and Goldstein, Daniel and Anthony, Quentin and Albalak, Alon and Alcaide, Eric and Biderman, Stella and Cheah, Eugene and Ferdinan, Teddy and Hou, Haowen and Kazienko, Przemys{\l}aw and others},
  journal={arXiv preprint arXiv:2404.05892},
  volume={3},
  year={2024}
}

@inproceedings{li2024ego,
  title={Is ego status all you need for open-loop end-to-end autonomous driving?},
  author={Li, Zhiqi and Yu, Zhiding and Lan, Shiyi and Li, Jiahan and Kautz, Jan and Lu, Tong and Alvarez, Jose M},
  booktitle={Proceedings of the IEEE/CVF Conference on Computer Vision and Pattern Recognition},
  pages={14864--14873},
  year={2024}
}

@inproceedings{yang2024parallelizing,
  title = {Parallelizing Linear Transformers with the Delta Rule over Sequence Length},
  author = {Yang, Songlin and Wang, Bailin and Zhang, Yu and Shen, Yikang and Kim, Yoon},
  booktitle = {Advances in Neural Information Processing Systems},
  pages = {115491--115522},
  year = {2024}
}

@article{yuan2024drama,
   title={Drama: An efficient end-to-end motion planner for autonomous driving with mamba},
   author={Yuan, Chengran and Zhang, Zhanqi and Sun, Jiawei and Sun, Shuo and Huang, Zefan and Lee, Christina Dao Wen and Li, Dongen and Han, Yuhang and Wong, Anthony and Tee, Keng Peng and others},
   journal={arXiv preprint arXiv:2408.03601},
   year={2024}
 }

@inproceedings{yang2024gated,
 author = {Yang, Songlin and Wang, Bailin and Shen, Yikang and Panda, Rameswar and Kim, Yoon},
 title = {Gated linear attention transformers with hardware-efficient training},
 year = {2024},
 booktitle = {Proceedings of the 41st International Conference on Machine Learning},
 pages = {23}
}

@article{li2024end,
  author={Chen, Li and Wu, Penghao and Chitta, Kashyap and Jaeger, Bernhard and Geiger, Andreas and Li, Hongyang},
  journal={IEEE Transactions on Pattern Analysis and Machine Intelligence}, 
  title={End-to-End Autonomous Driving: Challenges and Frontiers}, 
  year={2024},
  volume={46},
  number={12},
  pages={10164-10183}
}

@article{zhao2024survey,
  author={Zhao, Zhigen and Cheng, Shuo and Ding, Yan and Zhou, Ziyi and Zhang, Shiqi and Xu, Danfei and Zhao, Ye},
  journal={IEEE/ASME Transactions on Mechatronics},
  title={A Survey of Optimization-Based Task and Motion Planning: From Classical to Learning Approaches},
  year={2024},
  pages={1-27}
}

@article{pan2024safe,
  title={A safe motion planning and reliable control framework for autonomous vehicles},
  author={Pan, Huihui and Luo, Mao and Wang, Jue and Huang, Tenglong and Sun, Weichao},
  journal={IEEE transactions on intelligent vehicles},
  volume={9},
  number={4},
  pages={4780--4793},
  year={2024},
  publisher={IEEE}
}

@misc{yang2024fla,
  title = {FLA: A Triton-Based Library for Hardware-Efficient Implementations of Linear Attention Mechanism},
  author = {Yang, Songlin and Zhang, Yu},
  url = {https://github.com/fla-org/flash-linear-attention},
  year = {2024}
}

@inproceedings{wang2024llm,
  title={{Llm\textsuperscript{3}: Large language model-based task and motion planning with motion failure reasoning}},
  author={Wang, Shu and Han, Muzhi and Jiao, Ziyuan and Zhang, Zeyu and Wu, Ying Nian and Zhu, Song-Chun and Liu, Hangxin},
  booktitle={2024 IEEE/RSJ International Conference on Intelligent Robots and Systems (IROS)},
  pages={12086--12092},
  year={2024}
}

@article{jia2024bench2drive,
  title={Bench2drive: Towards multi-ability benchmarking of closed-loop end-to-end autonomous driving},
  author={Jia, Xiaosong and Yang, Zhenjie and Li, Qifeng and Zhang, Zhiyuan and Yan, Junchi},
  journal={Advances in Neural Information Processing Systems},
  volume={37},
  pages={819--844},
  year={2024}
}

@article{peng2025rwkv,
  title={Rwkv-7" goose" with expressive dynamic state evolution},
  author={Peng, Bo and Zhang, Ruichong and Goldstein, Daniel and Alcaide, Eric and Du, Xingjian and Hou, Haowen and Lin, Jiaju and Liu, Jiaxing and Lu, Janna and Merrill, William and others},
  journal={arXiv preprint arXiv:2503.14456},
  year={2025}
}

@article{li2025hydra,
  title={Hydra-MDP++: Advancing End-to-End Driving via Expert-Guided Hydra-Distillation},
  author={Li, Kailin and Li, Zhenxin and Lan, Shiyi and Xie, Yuan and Zhang, Zhizhong and Liu, Jiayi and Wu, Zuxuan and Yu, Zhiding and Alvarez, Jose M},
  journal={arXiv preprint arXiv:2503.12820},
  year={2025}
}

@article{jiang2025transdiffuser,
  title={TransDiffuser: End-to-end Trajectory Generation with Decorrelated Multi-modal Representation for Autonomous Driving},
  author={Jiang, Xuefeng and Ma, Yuan and Li, Pengxiang and Xu, Leimeng and Wen, Xin and Zhan, Kun and Xia, Zhongpu and Jia, Peng and Lang, XianPeng and Sun, Sheng},
  journal={arXiv preprint arXiv:2505.09315},
  year={2025}
}

@inproceedings{zheng2025genad,
  author={Zheng, Wenzhao and Song, Ruiqi and Guo, Xianda and Zhang, Chenming and Chen, Long},
  title={GenAD: Generative End-to-End Autonomous Driving},
  year={2025},
  publisher={Springer Nature Switzerland},
  booktitle={European Conference on Computer Vision},
  pages={87--104}
}

@article{guo2025ipad,
  title={iPad: Iterative Proposal-centric End-to-End Autonomous Driving},
  author={Guo, Ke and Liu, Haochen and Wu, Xiaojun and Pan, Jia and Lv, Chen},
  journal={arXiv preprint arXiv:2505.15111},
  year={2025}
}

@article{kimiteam2025kda,
  title={Kimi Linear: An Expressive, Efficient Attention Architecture}, 
  author={Kimi Team},
  year={2025},
  journal={arXiv preprint arXiv:2510.26692}
}

@inproceedings{jia2025drivetransformer,
  title={DriveTransformer: Unified Transformer for Scalable End-to-End Autonomous Driving},
  author={Xiaosong Jia and Junqi You and Zhiyuan Zhang and Junchi Yan},
  booktitle={International Conference on Learning Representations (ICLR)},
  year={2025}
}
